\documentclass[10pt,twocolumn,letterpaper]{article}

\usepackage[pagenumbers]{cvpr} 

\definecolor{cvprblue}{rgb}{0.21,0.49,0.74}
\usepackage[pagebackref,breaklinks,colorlinks,allcolors=cvprblue]{hyperref}

\usepackage{multirow}
\usepackage[most]{tcolorbox}
\tcbuselibrary{breakable}
\usepackage{colortbl}
\usepackage{pifont}
\usepackage{makecell}
\usepackage{cuted}
\usepackage[ruled,vlined,linesnumbered]{algorithm2e}

\usepackage{newfloat}
\usepackage{scrextend}
\usepackage{listings}
\usepackage{CJKutf8}

\def\paperID{} 
\def\confName{CVPR}
\def\confYear{2026}

\newcommand{\methodabbr}{QE}
\newcommand{\methodfull}{Quant Experts}

\title{\methodfull{}: Token-aware Adaptive Error Reconstruction with Mixture of Experts for Large Vision-Language Models Quantization}

\author{
Chenwei Jia\thanks{Equal contribution.} \quad
Baoting Li\footnotemark[1] \quad
Xuchong Zhang\thanks{Corresponding author: zhangxc0329@xjtu.edu.cn.} \quad
Mingzhuo Wei \quad
Bochen Lin \quad
Hongbin Sun\\
State Key Laboratory of Human-Machine Hybrid Augmented Intelligence\\
Institute of Artificial Intelligence and Robotics\\
Xi'an Jiaotong University \\
{\tt\small jiacw@stu.xjtu.edu.cn}
}

\begin{document}
\begin{CJK}{UTF8}{gbsn}

\maketitle
\begin{abstract}
Post-Training Quantization (PTQ) has emerged as an effective technique for alleviating the substantial computational and memory overheads of Vision-Language Models (VLMs) by compressing both weights and activations without retraining the full model.
Existing PTQ methods primarily rely on static identification and global compensation of sensitive or outlier channels, yet they often overlook the distributional differences of these important channels across inputs, leading to unsatisfactory quantization.
In this work, we observe that the distributions and occurrence frequencies of important channels vary significantly both across modalities and among tokens, even within the same modality. 
Accordingly, we propose \textbf{Quant Experts (QE)}, a token-aware adaptive error compensation with mixture-of-experts for VLMs quantization. 
QE divides the important channels into token-independent and token-dependent groups. 
For the former, a shared expert is designed for most tokens to compensate for global quantization error using a low-rank adapter. 
For the latter, routed experts including multiple routed low-rank adapters are elaborated to compensate for local quantization error related to specific tokens.
Extensive experiments demonstrate that QE consistently enhances task accuracy across various quantization settings and model scales, ranging from 2B to 70B parameters, while maintaining performance comparable to full-precision models.
\end{abstract}

\section{Introduction}\label{sec:intro}

Model quantization has become a key technique for reducing the computational and memory costs of large-scale multimodal models~\cite{kim2025efficient}. 
By mapping weights and activations to low-bit representations, it achieves substantial compression and acceleration while preserving accuracy. 
Among existing approaches, Post-Training Quantization (PTQ) is widely adopted for its efficiency and compatibility without retraining the full model. 
However, low-bit quantization inevitably introduces numerical perturbations that degrade performance, a problem that is especially pronounced in multimodal scenarios~\cite{2403.06408}.

\begin{figure}[t]
\centering
\includegraphics[width=\linewidth]{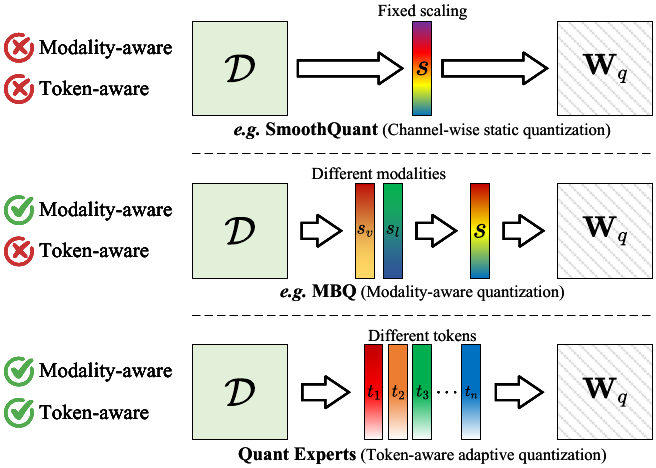}\label{fig:oracle}
\vspace{-15pt}
\caption{Illustration of Different Quantization Granularities. 
$\mathcal{D}$ is the calibration data, and $\mathbf{W}_q$ is the quantized weight matrix.
SmoothQuant~\cite{smoothquant} employs channel-wise with fixed scaling coefficients to achieve global quantization.
MBQ~\cite{Mbq} presents a modality-aware strategy that focuses on sensitivities in input modalities.
In contrast, we propose a token-aware adaptive quantization that considers both global and local error reconstruction.}
\vspace{-5pt}
\end{figure}

Recent studies on large-scale model quantization reveal significant variation in channel value distributions and sensitivities, where a small set of outlier or highly sensitive channels plays a key role in preserving model expressiveness and accuracy.
To mitigate quantization errors arising from these important channels, numerous methods have been proposed from different perspectives, including channel smoothing, mixed-precision strategy, Hessian-based optimization, and low-rank reconstruction approaches.
For instance, SmoothQuant~\cite{smoothquant} and AWQ~\cite{Awq} alleviate quantization errors by balancing activation and weight ranges via channel-wise scaling, with fixed coefficients for each channel estimated from the calibration dataset.
SpQR~\cite{spqr} statically identifies highly sensitive channels and allocates higher precision to them, effectively reducing quantization errors.
OBQ~\cite{obq} and GPTQ~\cite{gptq} estimate a static Hessian matrix from inputs and perform channel- or block-wise quantization based on the computed sensitivity.
LQER~\cite{lqer} and ASER~\cite{ASER} leverage singular value decomposition (SVD) to explicitly reconstruct quantization errors through a global low-rank adapter~\cite{lora}, \emph{i.e.}\ all important channels are handled uniformly by a single adapter.
Currently, in multimodal scenarios, MBQ~\cite{Mbq} reveals substantial differences in channel sensitivities across modalities and introduces a modality-aware channel scaling strategy to improve the adaptability of channel-smoothing methods.
Although these methods reduce quantization errors effectively, the fixed channel-importance estimation and global error compensation neglect the modeling of local feature variations, failing to adequately account for the dynamic relationships among important channel, error compensation, input modalities, and even tokens.

In this paper, we conduct a systematic analysis of large Vision-Language Models (VLMs) and observe that channel importance exhibits strong dynamics, varying significantly across modalities and tokens.
In particular, only a small subset of important channels appears consistently among most tokens, while the majority exhibit strong input-dependent fluctuations, with their importance differing across modalities and tokens.
To further minimize quantization errors, the dynamic nature of channel importance necessitates quantization methods that not only compensate for errors in globally activated important channels but also adaptively mitigate those arising in input-dependent ones.

Accordingly, we propose \textbf{\methodfull{} (\methodabbr)}, a token-aware adaptive error reconstruction framework with the Mixture-of-Experts (MoE)~\cite{moe,upcycling} for VLMs quantization, which considers both global and local variations in channel importance across different tokens.
\methodabbr{} first estimates the occurrence frequency of important channels from calibration data and then divides them into \emph{token-independent} and \emph{token-dependent} channels. 
We then employ two types of experts, the shared expert and the routed experts, to model these two kinds of channel groups respectively.
The shared expert employs a low-rank adapter to reconstruct global quantization errors that predominantly originate from token-independent important channels.
Meanwhile, token-dependent channels are clustered into multiple sub-groups according to their co-occurrence relationships, and each sub-group is assigned a routed expert equipped with a routed low-rank adapter.
During inference, \methodabbr{} fixedly employs the shared expert to compensate for global quantization errors and dynamically selects the most suitable routed expert according to the input tokens, thereby achieving adaptive compensation for local errors.
Extensive experiments under various quantization settings demonstrate that \methodabbr\ effectively mitigates errors caused by dynamic channel importance and substantially restores model performance under low-bit quantization.
This work makes the following contributions:
\begin{itemize}
\item We observe that the distributions and occurrence frequencies of important channels vary significantly both across modalities and among tokens, even within the same modality.
\item We propose \textbf{\methodabbr{}}, a token-aware adaptive error compensation with MoE for VLMs quantization. It employs a fixed shared expert to compensate global quantization errors while dynamically selecting routed experts to correct token-dependent local errors.
\item Extensive experiments across various quantization settings and multimodal benchmarks demonstrate that \methodabbr{} effectively mitigates low-bit performance degradation, achieving up to a 5.09\% accuracy improvement under the challenging W4A6 quantization for the 72B model.
\end{itemize}

\section{Observation and Motivation}\label{sec:mv}

\subsection{Observation One: The Positions of Important Channels Vary Across Tokens}\label{subsec:obs1}

\begin{figure}[t]
\centering
\includegraphics[width=1.0\linewidth]{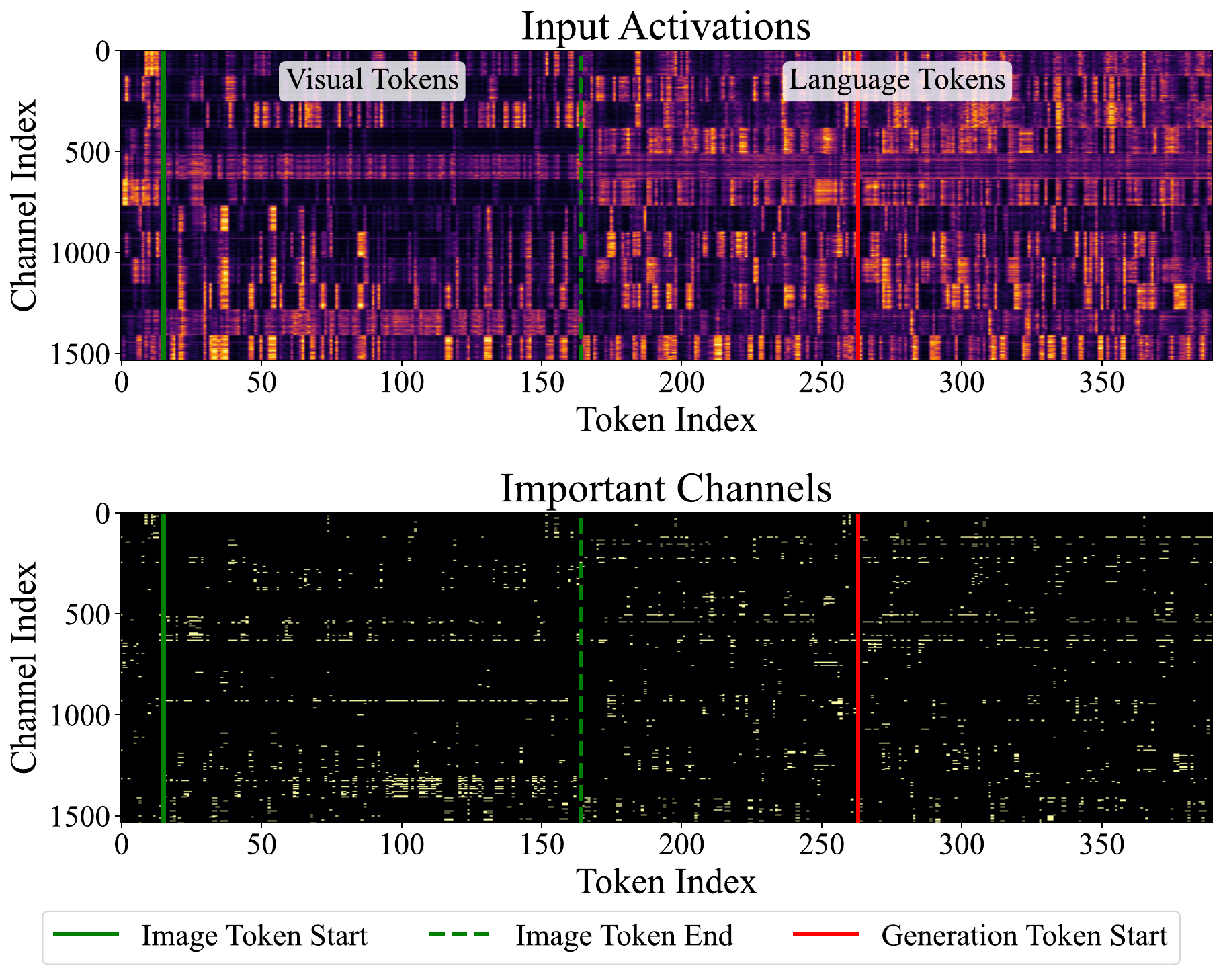}
\vspace{-20pt}
\caption{Visualization of a Transformer block in Qwen2VL-2B, illustrating token value distributions (top) and the positions of important channels (bottom). In the top panel, brightness reflects token magnitude, while in the bottom panel, highlighted positions denote important channels.}\label{fig:obs1}
\vspace{-5pt}
\end{figure}

As illustrated in \cref{fig:obs1}, we visualize the value distributions of several tokens along with the positions of their corresponding important channels computed by \cref{eq:impt1,eq:impt2}.
\begin{align}
\mathbf{w} :=& \operatorname{Mean}_{\text{row}}(|\mathbf{W}_f|),\label{eq:impt1}\\
\mathcal{C}_t =& \operatorname{Top}\text{-}k\!\left(|x_t| \odot \mathbf{w}\right), \label{eq:impt2}
\end{align}
where $\mathbf{W}_f \in \mathbb{R}^{d_\text{out}\times d_\text{in}}$ is the weight matrix, $d_\text{in}$ is the number of input channels, $d_\text{out}$ is the number of output channels, $\operatorname{Mean}_{\text{row}}$ computes the mean of absolute values along row, $\mathbf{w}$ is a $d_{\text{in}}$-dimensional vector, $x_t$ is the $t$-th token, and $\mathcal{C}_t$ denotes its top-$k$ important channels.

It is evident that, under the same layer weight, the positions of important channels vary not only across different modalities but also among tokens within the same modality.
In the case of across modalities, the intrinsic differences in value distributions lead to cross-modal shifts in the positions of important channels.
More importantly, even within the same modality, variations in token semantics and contextual information cause significant changes in activation distributions, resulting in dynamic migrations of important channels.
These findings indicate that the positions of important channels are not static but dynamically adapt to both inter- and intra-modality variations.

This observation reveals a key limitation of those globally calibrated and compensated methods~\cite{Awq,spqr,ASER}, just identifying some fixed outliers and employing unified compensation, which fails to capture token-wise variations in channel importance.

\subsection{Observation Two: Uneven Frequency Distribution of Important Channels}\label{subsec:obs2}

To gain deeper insight, we compute the occurrence frequency of each important channel using \cref{eq:freq} and sort them in descending order of frequency.
\begin{align}
f_c = k \times \frac{m_c}{\sum_{i=0}^{d_\text{in}} m_i},\label{eq:freq}
\end{align}
where $m_c$ denotes the number of times channel $c$ is identified as important, and $f_c$ represents the proportion of tokens for which channel $c$ is recognized as an important channel among all tokens.

\begin{figure}[t]
\centering
\includegraphics[width=1.0\linewidth]{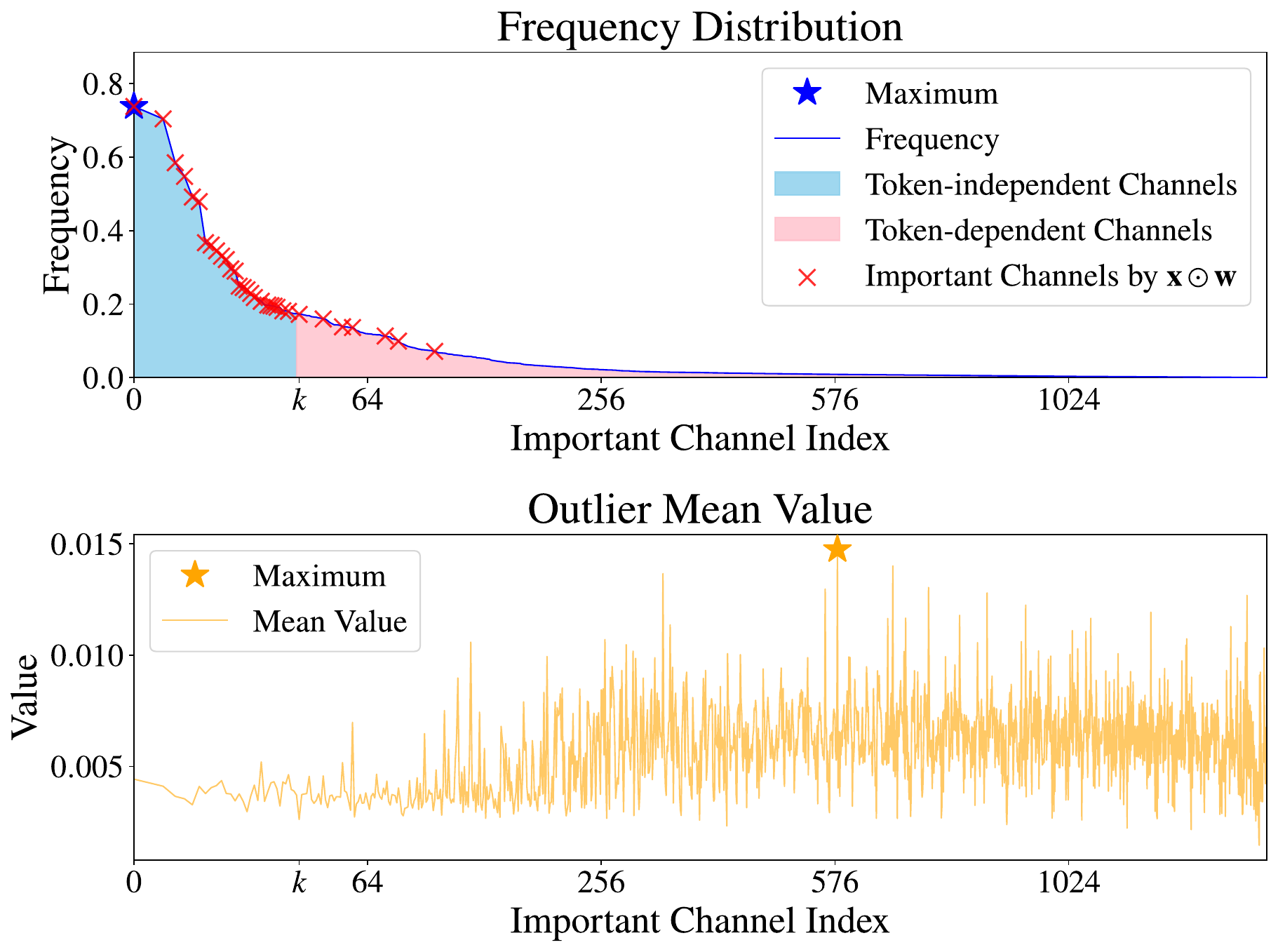}
\vspace{-20pt}
\caption{Visualization of a Transformer block in Qwen2VL-2B, illustrating important channel behavior. The top panel shows activation frequencies, while the bottom depicts average activation values. Red crosses mark important channel positions identified by the static global method on the calibration dataset.}
\label{fig:obs2}
\vspace{-5pt}
\end{figure}

As shown in~\cref{fig:obs2}, there is a small subset of important channels that occur across most tokens, while the majority are activated only for specific tokens. 
Meanwhile, those important channels with smaller frequency values may still exhibit larger outlier magnitudes, making their contribution to quantization error compensation equally non-negligible.
Based on this observation, we define the channels, which consistently appear across diverse input tokens and play a key role in compensating global quantization errors, as \emph{token-independent} important channels.
Conversely, channels whose activation depends on specific tokens are regarded as \emph{token-dependent} important channels, which reveal distinctive token properties of different tokens and the local nature of quantization errors.

This observation suggests that important channels in VLMs comprise both global and local components. 
Quantization error compensation that fails to properly address these two types inevitably results in performance degradation: insufficient compensation for token-independent important channels weakens the model’s overall representational capacity, while neglecting token-dependent important ones compromises token-level semantic fidelity. 
Therefore, an effective quantization framework should adopt differentiated protection and compensation strategies across tokens to ensure both stability and adaptability in performance.

\section{Method}\label{sec:method}

\begin{figure*}[t]
\centering
\includegraphics[width=\linewidth]{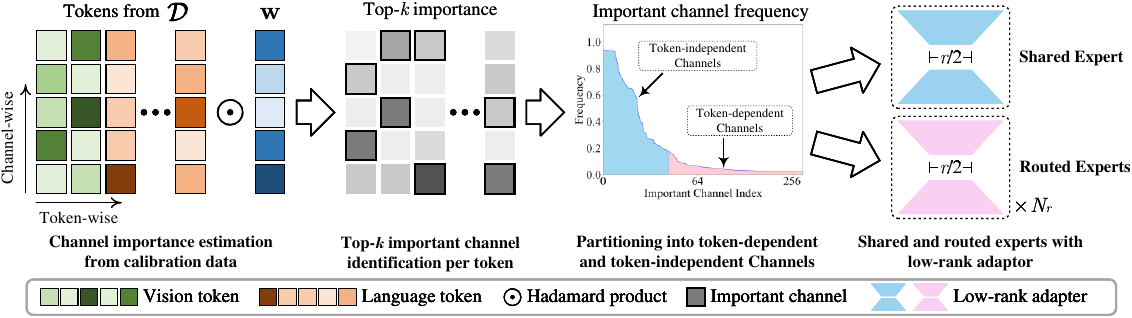}
\caption{The framework of \methodfull{} (\methodabbr{}). The token-independent channels are model by a shared expert, while token-dependent channels are captured by multiple routed experts. A lightweight low-rank adapter is implemented for each expert.}\label{fig:framework}
\end{figure*}

As illustrated in \cref{fig:framework}, \methodabbr{} first estimates the distribution frequency of important channels from the calibration data $\mathcal{D}$ and partitions them into \emph{token-independent} and \emph{token-dependent} channels. 
It then employs the \textbf{\emph{shared expert (SE)}} and the \textbf{\emph{routed experts (REs)}}, to handle these distinct channel groups, each implemented as a low-rank adapter.
The shared expert focuses on global quantization error that is mainly caused by token-independent channels.
Whereas token-dependent channels are clustered through co-occurrence analysis into multiple subgroups, and each is handled by a routed expert that specializes in local error compensation.
During inference, a lightweight router dynamically selects the optimal routed expert for adaptive error compensation.

\subsection{Calibration and Error Reconstruction}
In this subsection, we illustrate how to minimize quantization error from token-aware adaptive optimization and establish the mapping between important channels and the mixture-of-experts.
Let $\mathbf{W}_f^l$ denote the full-precision weight of the $l$-th layer, and $\mathbf{W}_q^l\!=\!Q(\mathbf{W}_f^l)$ its quantized counterpart. 
As typical, the quantization error is defined as $\mathbf{E}^l\!=\!\mathbf{W}_f^l\!-\!\mathbf{W}_q^l$.
Building on the analysis in \cref{sec:mv}, we further formulate quantization objective optimization as:
\begin{align}
\arg\min_{\tilde{\mathbf{E}}^l} \| (\mathbf{E}^l - \underbrace{(\tilde{\mathbf{E}}^l_S + \tilde{\mathbf{E}}^l_R(x^l))}_{\tilde{\mathbf{E}}^l})x^l \|_F,\label{eq:qe_obj}
\end{align}
where $x^l$ denotes the input tokens of the $l$-th layer, $\tilde{\mathbf{E}}^l_S$ represents the token-independent error approximation, and $\mathbf{E}^l_R(x^l)$ denotes the error approximation dependent on $x^l$.

Our approach leverages two complementary low-rank adapters to reconstruct the quantization error:
\begin{align}
\tilde{\mathbf{E}}^l 
&= 
\underbrace{\mathbf{L}_{SA}^l \mathbf{L}_{SB}^l}_{\tilde{\mathbf{E}}^l_S} 
+ 
\underbrace{\mathbf{L}_{RA}^{l,i^*} \mathbf{L}_{RB}^{l,i^*}}_{{\mathbf{E}}^l_R(x^l)},\label{eq:qe_error} \\
i^* &= \arg\min_i \left( \mathbf{R}^l |x^l| \right)_i,
\end{align}
where $\mathbf{L}_{SA}^l$ and $\mathbf{L}_{SB}^l$ form the shared expert that models token-independent error approximation, while $\mathbf{L}_{RA}^{l,i}$ and $\mathbf{L}_{RB}^{l,i}$ denote the routed experts adaptively modeling token-dependent error approximation. 
$\mathbf{R}^l$ is a lightweight router that predicts a score for each routed expert, and selects an expert with the minimal score. 
In this way, \methodabbr{} integrates global token-independent and local token-dependent compensation within a unified mixture-of-experts framework.

We quantify the distribution of channel importance using activation $X\!\in\!\mathbb{R}^{T\times{d_{in}}}$ collected from the calibration data $\mathcal{D}$.
For the $t$-th input token $x_t^l$ in layer $l$, we compute the per-token important channels $\mathcal{A}^l_t$ as \cref{eq:impt2}.
Then we aggregate all $\mathcal{A}^l_t$ over $\mathcal{D}$ to form a multiset $\mathcal{T}^l$, and compute the occurrence frequency $f^l$ of each important channel according to \cref{eq:freq}.
These channels are sorted by $f$ and partitioned into two disjoint important channel sets: 
the first $k$ channels $\mathcal{C}^l_s$, referred to as token-independent channels, 
and the subsequent $(N_r k)$ channels $\mathcal{C}^l_r$, referred to as token-dependent channels.
To sum up, the pseudo-code of the dependence partitioning process is detailed in \cref{alg:chan-partition}.

\begin{algorithm}[t]
\caption{Channel Dependence Partitioning}
\label{alg:chan-partition}
\DontPrintSemicolon
\SetKwFunction{TopK}{TopK}
\SetKwFunction{Argsort}{argsort}
\SetKwFunction{Unique}{unique}
\SetKwInOut{Input}{Input}\SetKwInOut{Output}{Output}
\Input{Layer weight $\{\mathbf{W}^{l}_f\}_{l=1}^{L}$; calibration data $\mathcal{D}$; the number of important channels $k$; the number of routed experts $N_r$.}
\Output{Token-independent channels $\{\mathcal{C}^l_s\}$; token-dependent channels $\{\mathcal{C}^l_r\}$.}
\BlankLine

\For{sample $d\in\mathcal{D}$}{
  Inference $d$ and store input activations in $\mathbf{X}$\;
  }
\For{$l\gets1$ \KwTo $L$}{
Initialize $\mathcal{T}^l\leftarrow\emptyset$\;
Compute $\mathbf{w}^l$ from $\mathbf{W}^l_f$ according to \cref{eq:impt1}\;
  \For{token $x_t^l$ in $\mathbf{X}^{l}$}{
    $\mathcal{A}^l_t \leftarrow \operatorname{Top}\text{-}k(|x^{l}_t| \cdot \mathbf{w}^l)$ \tcp*{indices}
    $\mathcal{T}^l \leftarrow \mathcal{T}^l\uplus \mathcal{A}^l_t$\;
  }
  $(\mathcal{C}^l_{\text{unsorted}},f^l)\leftarrow \Unique(\mathcal{I}^l)$\;
  $\pi\leftarrow \Argsort(f^l,\mathrm{descending})$\;
  Reorder $\mathcal{C}^l\leftarrow \mathcal{C}^l_{\text{unsorted}}[\pi]$\;
  Token-independent channels $\mathcal{C}^l_s \leftarrow \mathcal{C}^l_{[1{:}k]}$\;
  Token-dependent channels $\mathcal{C}^l_r \leftarrow \mathcal{C}^l_{[k+1{:}(N_r+1)k]}$
}
\Return{$\{\mathcal{C}^l_s\}_{l=1}^{L},\ \{\mathcal{C}^l_r\}_{l=1}^{L}$}\;
\end{algorithm}

\subsection{\textbf{\emph{SE}} for Token-Independent Channels}

We introduce the shared expert to reconstruct the quantization error that primarily arises from token-independent channels.
As in prior work~\cite{Awq,gptq,easyquant}, the performance degradation and quantization error stem mainly from these outlier channels, underscoring the need to mitigate their impact on both weight and activation quantization.

Building upon the above motivation, the shared expert is designed following~\cite{ASER} to focus on the token-independent channels $\mathcal{C}^l_s$.
Specifically, to alleviate weight quantization errors and suppress the interference of their outlier magnitudes on other channels, token-independent channels are exempted from direct quantization and reconstructed by a low-rank adapter $\{(\mathbf{L}_{SA}^l,\mathbf{L}_{SB}^l)\}^L_l$ using whitening SVD. 
To mitigate token-wise activation quantization errors, the shared expert employs channel-wise scaling, which reduces activation magnitudes while proportionally amplifying the corresponding weights.
Through the initial reconstruction for $\mathbf{E}^l$ performed by the shared expert, \methodabbr{} achieves precise recovery of token-independent channels, while the remaining error $\mathbf{E}_S^l\!=\!\mathbf{E}^l\!-\!\mathbf{L}_{SA}^l \mathbf{L}_{SB}^l$ is subsequently refined by the routed experts associated with token-dependent channels.
The algorithm for the shared expert is detailed in the supplementary~\cref{subsec:se}.

\subsection{\textbf{\emph{REs}} for Token-Dependent Channels}\label{sec:method:res}

\begin{algorithm}[t]
\caption{Building Routed Experts}
\label{alg:routed-expert}
\DontPrintSemicolon
\SetKwInOut{Input}{Input}
\SetKwInOut{Output}{Output}
\Input{Per-layer data $\{\!\mathbf{X}^l,\,\mathcal{O}^l,\,\mathbf{E}_S^l,\,\mathcal{C}^l_r\!\}_{l=1}^{L}$; the number of routed experts $N_r$; rank $r$.}
\Output{
  REs $\{(\mathbf{L}_{RA}^{l,i},\, \mathbf{L}_{RB}^{l,i})\}$ and Router $\{\mathbf{R}^{l}\}$.
}
\BlankLine
\For{$l\gets1$ \KwTo $L$}{
Initialize router parameter $\mathbf{R}^l \in \mathbb{R}^{d_{\text{in}} \times N_r}$\;
Compute $\mathbf{x}^l \leftarrow \operatorname{Mean}_{\text{row}}(|\mathbf{X}^l|)$\;
$\mathbf{S}^l \leftarrow \operatorname{NPMI\_Similarity}(\mathcal{O}^l)$\;
$\mathbf{\mathcal{N}}^l \leftarrow \operatorname{Normed\_Laplacian}(\mathbf{S}^l)$\;
$\mathbf{U}^l \leftarrow \operatorname{Eigenvectors}(\mathbf{\mathcal{N}}^l)$\;
Cluster $\mathbf{\Gamma}^l \leftarrow \operatorname{KMeans}(\mathbf{U}_{[:,2:N_r+1]}^l,\ N_r)$\;
\For{cluster labels $\gamma\in\mathbf{\Gamma}^l$}{
    Initialize weight $\omega \leftarrow \mathbf{1} \in \mathbb{R}^{d_{\text{in}}}$\;
    Compute $\omega \leftarrow ({\mathbf{x}^l_{\gamma}}/{\min(\mathbf{x}^l_{\gamma})}) \odot \mathbf{x}^l$\;
    Normalize $\omega \leftarrow {\omega} / {\sqrt{\min(\omega)\max(\omega)}}$\;
    Perform SVD: $U\Sigma V^\top = \mathbf{E}_S^l\mathrm{diag}(\omega)$\;
    $\mathbf{L}_{RA}^{l,i} \leftarrow U_r \Sigma_r$\;
    $\mathbf{L}_{RB}^{l,i} \leftarrow V_r^\top S^{-1} \mathrm{diag}(1/\omega)$\;
    $R^l_{[:,i]} \leftarrow \operatorname{Mean}_{\text{row}}(|\mathbf{E}_S^l - \mathbf{L}_{RA}^{l,i} \mathbf{L}_{RB}^{l,i}|)$\;
}
}
\Return{$\{(\mathbf{L}_{RA}^{l,i},\, \mathbf{L}_{RB}^{l,i})\}_{l=1,i=1}^{L,N_r}$, $\{\mathbf{R}^{l}\}_{l=1}^{L}$}\;
\end{algorithm}

In this subsection, we focus on the quantization error compensation for token-dependent important channels.
Ideally, one would tailor an individual compensation strategy to each token, but the virtually unbounded value combinations make such token-specific designs computationally infeasible.
Therefore, it becomes essential to design a constrained yet effective compensation strategy that approximates the optimal solution.
We observe that token-dependent channels exhibit correlated occurrence patterns across different tokens.
Based on this insight, we empirically compute their co-occurrence statistics and cluster channels with strong mutual association. 
We then employ routed experts, each employed with a low-rank adapter dedicated to modeling its corresponding channel cluster.
During inference, a lightweight router estimates the final error of each expert and activates the one predicted to yield the lowest error.

Specifically, we construct the co-occurrence matrix following~\cref{eq:occ} to capture the token-level correlation patterns among token-dependent channels from all tokens $\mathbf{X}^l$.
\begin{align}
\mathcal{O}^l_{t,i} = \mathbf{1}\big(c_i \in \mathcal{C}^l_r \cap \mathcal{A}^l_t \big), \quad \mathcal{O}^l\!\in\!\{0,1\}^{T \times (N_r k)}, \label{eq:occ}
\end{align}
where $\mathcal{O}^l$ denotes the co-occurrence matrix over $T$ tokens in $l$-th layer, and the indicator function $\mathbf{1}(\cdot)$ returns $1$ if the $i$-th channel $c_i$ is important for the $t$-th token, and $0$ otherwise.

Next, we employ spectral clustering~\cite{normcuts, spectral} to partition the token-dependent important channels based on their most likely co-occurrence patterns.
Firstly, the co-occurrence matrix $\mathcal{O}^l$ is transformed into a similarity matrix $\mathbf{S}^l\!\in\!\mathbb{R}^{(N_r k)\!\times\!(N_r k)}$ by using the normalized pointwise mutual information (NPMI)~\cite{pmi, npmi}, which quantifies the association strength between co-occurring channels as follows:
\begin{align}
p(i) &= \tfrac{1}{T}\sum_t^T \mathcal{O}^l_{t,i}, \\
p(i,j) &= \tfrac{1}{T}\sum_t^T (\mathcal{O}^l_{t,i} \mathcal{O}^l_{t,j}), \\
\mathbf{S}_{i,j} &= (\log \tfrac{p(i,j)}{p(i)p(j)})\big/-\log p(i,j),
\label{eq:npmi}
\end{align}
where $i, j$ represent the channel indices, and $\mathbf{S}_{i,j}$ represents the NPMI value between channels $i$ and $j$. 
A higher NPMI value indicates a stronger likelihood that these channels will occur as important channels within a single token.

\begin{figure}[t]
\centering
\includegraphics[width=0.8\linewidth]{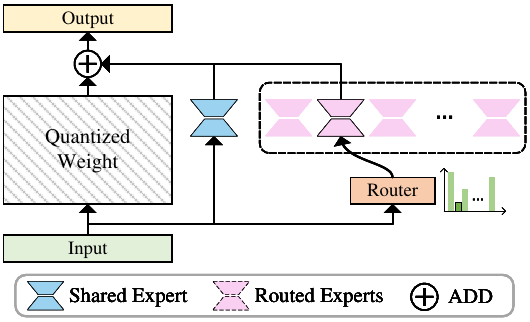}
\vspace{-5pt}
\caption{Illustration of the Inference Computation Process of~\methodabbr{}.}
\label{fig:infer}
\end{figure}

For the similarity matrix $\mathbf{S}^l$, we perform eigen decomposition on its normalized Laplacian $\mathbf{\mathcal{N}}^l$ and take the next $N_r$ eigenvectors $\mathbf{U}^l\!\!=\![u_2,\!\ldots,\!u_{N_r+1}]$ as spectral embeddings. 
K-Means then partitions $\mathcal{C}^l_r$ into $N_r$ clusters of token-dependent channels using spectral embeddings $\mathbf{U}^l$.
For each cluster, we apply a weighting vector $\omega$ on $\mathbf{E}_S^l$ to enhance the reconstruction accuracy for its token-dependent channels $\gamma$.
Next, we perform SVD to reconstruct each weighted $\mathbf{E}_S^l$ and truncate the rank to $r$.
The $N_r$ low-rank adapters together constitute the routed experts.
Finally, we use the absolute mean of the remaining error $\mathbf{E}_R^{l,i}\!=\!\mathbf{E}_S^l\!-\!\mathbf{L}_{RA}^{l,i} \mathbf{L}_{RB}^{l,i}$ for $i$-th routed expert as the parameters of router $\mathbf{R}^{l}_i$, which estimates the error of any input token under $i$-th routed expert.
The above procedure is detailed in~\cref{alg:routed-expert}.
The inference process is illustrated in~\cref{fig:infer}, where the shared expert provides stable global compensation and the router adaptively activates the optimal routed expert for dynamic local compensation.

To further alleviate the performance degradation introduced by post-training quantization, we design an optional lightweight refinement strategy.
Specifically, only routed experts $(\mathbf{L}_{RA}^{l}, \mathbf{L}_{RB}^{l})$ and the router $\mathbf{R}^{l}$ are trainable, while all other parameters remain frozen.
Moreover, this refinement is performed layer-wise without end-to-end training of all parameters.
The refinement is detailed and formulated in the supplementary material~\cref{suppl:ft}.
\begin{table*}[t]
  \centering
  \resizebox{1.0\textwidth}{!}{
      \begin{tabular}{l|cc|ccccccccccc|c}
        \toprule
        Method & \#W & \#A & MMMU & OCRBench & ScienceQA & TextVQA & VizWiz & AI2D & ChartQA & DocVQA & InfoVQA & MMStar & MuriBench & Avg. ($\uparrow$) \\
        \midrule
    
        Qwen2VL-2B & 16 & 16 & 39.89 & 74.90 & 76.96 & 77.72 & 65.73 & 70.01 & 72.04 & 87.28 & 58.49 & 43.46 & 26.19 & 62.97 \\
        \midrule
        \midrule
        
        RTN & 4 & 6 & 34.00 & 59.80 & 64.70 & 67.58 & 55.62 & 59.26 & 56.08 & 75.78 & 45.12 & 40.67 & 31.19 & 53.62\\
        SQ (ICML'23) & 4 & 6 & 30.44 & 59.60 & 65.25 & 65.88 & 53.90 & 59.16 & 40.44 & 70.73 & 39.36 & 38.20 & 30.00 & 50.27\\
        LQER (ICML'24) & 4 & 6 & 33.00 & 65.80 & 68.32 & 69.37 & 55.91 & 62.56 & \textbf{62.68} & 81.02 & 48.92 & 37.82 & 29.77 & 55.92\\
        MBQ (CVPR'25) & 4 & 6 & \textbf{34.44} & 61.10 & 67.08 & 69.45 & 57.19 & 60.91 & 60.08 & 76.24 & 43.13 & \textbf{42.61} & 29.77 & 54.73\\
        \midrule
        \methodabbr{} & 4 & 6 & 33.78 & \textbf{68.20} & \textbf{71.84} & \textbf{73.18} & \textbf{59.62} & \textbf{65.45} & 64.60 & \textbf{82.75} & \textbf{51.84} & 42.04 & \textbf{32.88} & \textbf{58.74}\\

        \midrule
        \midrule
        
        RTN & 4 & 8 & 35.00 & 65.20 & 70.85 & 72.46 & 56.96 & 65.45 & 65.24 & 80.10 & 49.03 & 40.75 & \textbf{34.23} & 57.75\\
        SQ (ICML'23) & 4 & 8 & 32.11 & 65.80 & 68.02 & 68.61 & 57.43 & 61.95 & 44.08 & 75.07 & 42.06 & 38.93 & 29.58 & 53.06\\
        LQER (ICML'24) & 4 & 8 & 35.67 & 69.80 & 71.99 & 73.08 & 56.81 & 67.55 & 66.68 & 83.68 & 52.73 & 39.88 & 31.50 & 59.03\\
        MBQ (CVPR'25) & 4 & 8 & 34.33 & 62.30 & 70.85 & 72.36 & 58.32 & 65.19 & 62.32 & 78.19 & 47.93 & \textbf{42.57} & 32.62 & 57.00\\
        \midrule
        \methodabbr{} & 4 & 8 & \textbf{37.33} & \textbf{72.10} & \textbf{74.67} & \textbf{75.34} & \textbf{61.59} & \textbf{68.36} & \textbf{69.28} & \textbf{84.46} & \textbf{53.97} & 41.29 & 34.12 & \textbf{61.14}\\

        \midrule
        \midrule
        
        RTN & 3 & 16 & 32.44 & 65.80 & 67.63 & 70.43 & 55.81 & 61.59 & 62.80 & 78.36 & 46.32 & 35.24 & 38.54 & 55.91\\
        AWQ (MLSys'24) & 3 & 16 & 33.22 & 63.40 & 68.32 & 70.60 & 56.88 & 61.72 & 62.08 & 78.70 & 45.02 & 36.48 & 35.62 & 55.64\\
        LQER (ICML'24) & 3 & 16 & \textbf{34.56} & 67.50 & 69.86 & 70.04 & 58.68 & 62.76 & 65.36 & 81.19 & 47.54 & 36.35 & \textbf{38.58} & 57.49\\
        MBQ (CVPR'25) & 3 & 16 & 33.44 & 63.90 & 69.16 & 70.75 & 52.59 & 63.28 & 63.56 & 79.27 & 45.87 & 34.66 & 34.50 & 55.54\\
        \midrule
        \methodabbr{} & 3 & 16 & 33.89 & \textbf{70.10} & \textbf{72.09} & \textbf{74.46} & \textbf{60.15} & \textbf{64.96} & \textbf{68.52} & \textbf{82.97} & \textbf{51.31} & \textbf{39.66} & 34.12 & \textbf{59.29}\\

        \bottomrule
      \end{tabular}
    }
    \vspace{-5pt}
  \caption{Main results on the model of Qwen2VL-2B.}\label{tab:qwen2b}
  \vspace{-5pt}
\end{table*}

\begin{table*}[t]
  \centering
  \resizebox{1.0\textwidth}{!}{
      \begin{tabular}{l|cc|ccccccccccc|c}
        \toprule
        Method & \#W & \#A & MMMU & OCRBench & ScienceQA & TextVQA & VizWiz & AI2D & ChartQA & DocVQA & InfoVQA & MMStar & MuriBench & Avg. ($\uparrow$) \\
        \midrule
    
        InternVL2-8B & 16 & 16 & 48.00 & 76.90 & 97.12 & 76.91 & 60.61 & 82.09 & 82.60 & 89.97 & 66.92 & 59.36 & 36.12 & 70.60 \\
        \midrule
        \midrule
        
        RTN & 4 & 6 & 37.00 & 69.20 & 93.16 & 69.96 & 55.34 & 73.77 & 74.16 & 83.81 & 55.82 & 50.03 & 30.12 & 62.94\\
        SQ (ICML'23) & 4 & 6 & 40.44 & 69.50 & 94.84 & 69.83 & 50.91 & 75.13 & 74.92 & 84.28 & 56.77 & 49.59 & 32.00 & 63.47\\
        LQER (ICML'24) & 4 & 6 & 40.22 & 72.20 & 94.74 & 71.83 & 58.03 & 76.49 & 77.88 & 85.80 & 60.47 & 49.69 & 30.85 & 65.29\\
        MBQ (CVPR'25) & 4 & 6 & 43.67 & 71.00 & 95.49 & 70.26 & 52.90 & 77.59 & 75.64 & 84.21 & 58.09 & 53.89 & 32.27 & 65.00\\
        \midrule
        \methodabbr{} & 4 & 6 & \textbf{44.89} & \textbf{74.30} & \textbf{96.23} & \textbf{74.77} & \textbf{59.00} & \textbf{79.40} & \textbf{80.48} & \textbf{87.77} & \textbf{63.12} & \textbf{55.27} & \textbf{34.15} & \textbf{68.13}\\

        \midrule
        \midrule
        
        RTN & 4 & 8 & 43.33 & 72.80 & 95.93 & 73.11 & 56.96 & 79.40 & 79.36 & 86.55 & 61.47 & 55.23 & \textbf{35.38} & 67.23\\
        SQ (ICML'23) & 4 & 8 & 42.22 & 72.20 & 95.54 & 72.58 & 51.68 & 77.04 & 77.48 & 85.48 & 59.07 & 51.98 & 32.58 & 65.26\\
        LQER (ICML'24) & 4 & 8 & 44.44 & 75.10 & 96.68 & 75.06 & 57.49 & 80.99 & 80.44 & 88.11 & 63.85 & 54.77 & 33.15 & 68.19\\
        MBQ (CVPR'25) & 4 & 8 & 44.44 & 73.50 & \textbf{96.78} & 72.00 & 56.67 & 79.21 & 77.72 & 86.42 & 61.48 & 55.32 & 32.77 & 66.94\\
        \midrule
        \methodabbr{} & 4 & 8 & \textbf{45.56} & \textbf{75.60} & 96.73 & \textbf{76.08} & \textbf{59.77} & \textbf{81.06} & \textbf{81.60} & \textbf{88.31} & \textbf{64.83} & \textbf{56.09} & 34.31 & \textbf{69.09}\\

        \midrule
        \midrule
        
        RTN & 3 & 16 & 44.22 & 74.20 & 96.18 & 74.64 & 55.99 & 80.47 & 79.32 & 87.96 & 62.61 & 55.37 & 32.88 & 67.62\\
        AWQ (MLSys'24) & 3 & 16 & 45.67 & 74.60 & 96.33 & 74.97 & 59.18 & 80.47 & 80.08 & 88.01 & 63.55 & 54.55 & 34.85 & 68.39\\
        LQER (ICML'24) & 3 & 16 & 45.33 & 74.90 & \textbf{96.38} & 74.77 & 57.12 & 80.60 & 80.04 & 88.10 & 63.61 & 55.65 & 32.73 & 68.11\\
        MBQ (CVPR'25) & 3 & 16 & \textbf{46.11} & 75.20 & 96.18 & 74.97 & 58.43 & 79.70 & 79.72 & 88.00 & 63.16 & 54.96 & \textbf{35.00} & 68.31\\
        \midrule
        \methodabbr{} & 3 & 16 & 45.78 & \textbf{75.90} & 96.28 & \textbf{75.50} & \textbf{59.46} & \textbf{80.99} & \textbf{80.76} & \textbf{88.79} & \textbf{64.35} & \textbf{57.14} & 33.38 & \textbf{68.94}\\

        \bottomrule
      \end{tabular}
    }
    \vspace{-5pt}
  \caption{Main results on the model of InternVL2-8B.}\label{tab:internvl8b}
  \vspace{-5pt}
\end{table*}

\section{Experiments}\label{sec:exp}

\subsection{Experimental Setup}
\noindent\textbf{Models.} 
We conduct comprehensive PTQ experiments on several representative open-source VLMs, covering the Qwen2VL~\cite{qwen2} series (2B, 7B, 72B) and the InternVL2~\cite{internvl} series (2B, 8B).
All model weights are obtained from the official repositories.

\noindent\textbf{Baselines.} 
We perform systematic comparisons with popular open-source PTQ methods.
For weight-activation quantization, \methodabbr{} is evaluated under W4A6 and W4A8 settings against round-to-nearest (RTN), channel-scaling SmoothQuant (SQ)~\cite{smoothquant}, modality-balanced MBQ~\cite{Mbq}, and low-rank reconstruction LQER~\cite{lqer}.
Activations and weights are quantized using per-token and per-output-channel symmetric schemes, respectively.
For weight-only quantization, we adopt AWQ~\cite{Awq} as the channel-scaling baseline under the W3A16 configuration, applying group-wise asymmetric quantization with 128 group size.
Throughout, ``W$x$A$y$'' denotes weight and activation bitwidths of $x$ and $y$, while \#W and \#A represent their bitwidths respectively.

\noindent\textbf{Evaluation Metrics.} 
To comprehensively assess the performance of our method, we experiment across diverse multimodal tasks.
Text recognition and understanding are tested on OCRBench~\cite{ocrbench} and TextVQA~\cite{textvqa}; 
document and infographic comprehension on DocVQA~\cite{docvqa} and InfoVQA~\cite{infographicvqa}; 
chart reasoning on ChartQA~\cite{chartqa}; and general visual perception on VizWiz-VQA~\cite{vizwiz}.
ScienceQA~\cite{scienceqa} and MMMU~\cite{mmmu} evaluate scientific and general reasoning, while MMStar~\cite{mmstar} and MuirBench~\cite{muirbench} assess overall multimodal and multi-image understanding.
AI2D~\cite{ai2d} measures diagram comprehension, ensuring coverage across all major aspects of multimodal reasoning.
We evaluate on the open source evaluation framework LMMs-Eval~\cite{lmms}.

\noindent\textbf{Experimental Details.} 
We follow MBQ~\cite{Mbq} and use the enhanced COCO Caption dataset~\cite{coco} from ShareGPT4V~\cite{sharegpt4v}, randomly sampling 128 image-caption pairs as the calibration set. 
In both LQER and \methodabbr{}, the total SVD rank $r$ is set to 64. 
Since \methodabbr{} employs both shared and routed experts, the total rank of 64 is split into $\frac{64}{2}$ for each type, ensuring that the overall rank matches LQER.
The $k$ is fixed to 32, and the $N_r$ is set to 8.
For refinement, we set epochs for 16 with 100 iterations per epoch using the AdamW~\cite{adamw} optimizer (learning rate $1\times10^{-4}$, no weight decay) and a cosine annealing schedule~\cite{cos}.
The refinement coefficients are $\tau = 0.5$, $\alpha = 1.0$, and $\beta = 0.05$.
Smaller models are evaluated on $4\times$ RTX~4090 24G GPUs, and the 72B model is on $4\times$ A800 80G GPUs.

\subsection{Main Results}

Smaller VLMs are generally more sensitive to quantization, making them more challenging benchmarks~\cite{li2024evaluating,Mbq}.
We report results on Qwen2VL-2B and InternVL2-8B in \cref{tab:qwen2b} and \cref{tab:internvl8b}, and on the larger 72B model in \cref{tab:qwen72b}.

\noindent\textbf{Weight-Activation Quantization.} 
\methodabbr{} consistently surpasses both the modality-aware baseline MBQ~\cite{Mbq} and the static low-rank method LQER~\cite{lqer}.
In the challenging W4A6 setting, it improves Qwen2VL-2B accuracy by 4.01\% over MBQ, with only a 4.23\% drop from full precision, and gains 3.13\% on InternVL2-2B.
At W4A8, the performance gap to full precision narrows to within 2\%, demonstrating strong robustness.
On the larger Qwen2VL-72B, \methodabbr{} achieves a remarkable average accuracy improvement of 5.09\% under the W4A6 quantization setting, nearly matching full-precision performance.

\noindent\textbf{Weight-Only Quantization.} 
\methodabbr{} again outperforms MBQ and LQER across all models.
MBQ’s distribution reshaping offers limited benefit over AWQ due to capacity limits, while LQER’s static compensation yields only minor recovery.
These results highlight the necessity of dynamic compensation to handle modality- and token-level distribution shifts that static methods cannot capture.

Additional results on Qwen2VL-7B and InternVL2-2B (Supplementary \cref{tab:qwen7b} and \cref{tab:internvl2b}) show consistent gains, further confirming the generalizability of our method across diverse vision-language models.

\begin{table}[h]
\centering
\resizebox{\linewidth}{!}{
\begin{tabular}{c|l|cccccc}
\toprule
Setting & Method & MMMU & OCRBench & ScienceQA & TextVQA & VizWiz \\
\midrule
FP16 & - & 61.44 & 78.70 & 91.22 & 82.26 & 76.27 \\
\midrule
\multirow{5}{*}{W4A6} 
& RTN  & 51.33 & 59.10 & 85.52 & 72.12 & 64.29 \\
& SQ   & 53.33 & 61.20 & 86.96 & 74.42 & 63.45 \\
& LQER & 52.33 & 59.60 & 86.71 & 73.84 & 66.64 \\
& MBQ  & 52.67 & 69.70 & 86.32 & 76.08 & 67.99 \\
\cmidrule(lr){2-7}
& \methodabbr\ & \textbf{58.11} & \textbf{76.60} & \textbf{90.33} & \textbf{79.27} & \textbf{73.91} \\
\midrule
\multirow{5}{*}{W4A8} 
& RTN  & 57.11 & 66.50 & 90.08 & 75.96 & 71.14 \\
& SQ   & 56.44 & 65.10 & 90.03 & 76.37 & 66.70 \\
& LQER & 57.44 & 70.60 & 91.32 & 78.15 & 74.28 \\
& MBQ  & 58.33 & 73.90 & 89.24 & 79.19 & 72.24 \\
\cmidrule(lr){2-7}
& \methodabbr\ & \textbf{58.89} & \textbf{78.30} & \textbf{91.47} & \textbf{81.47} & \textbf{75.83} \\
\bottomrule
\end{tabular}
}
\vspace{-5pt}
\caption{Main results of Qwen2VL-72B model (higher is better).}\label{tab:qwen72b}
\vspace{-5pt}
\end{table}

\subsection{Ablation Studies}

\begin{figure*}[t]
\centering
\includegraphics[width=\linewidth]{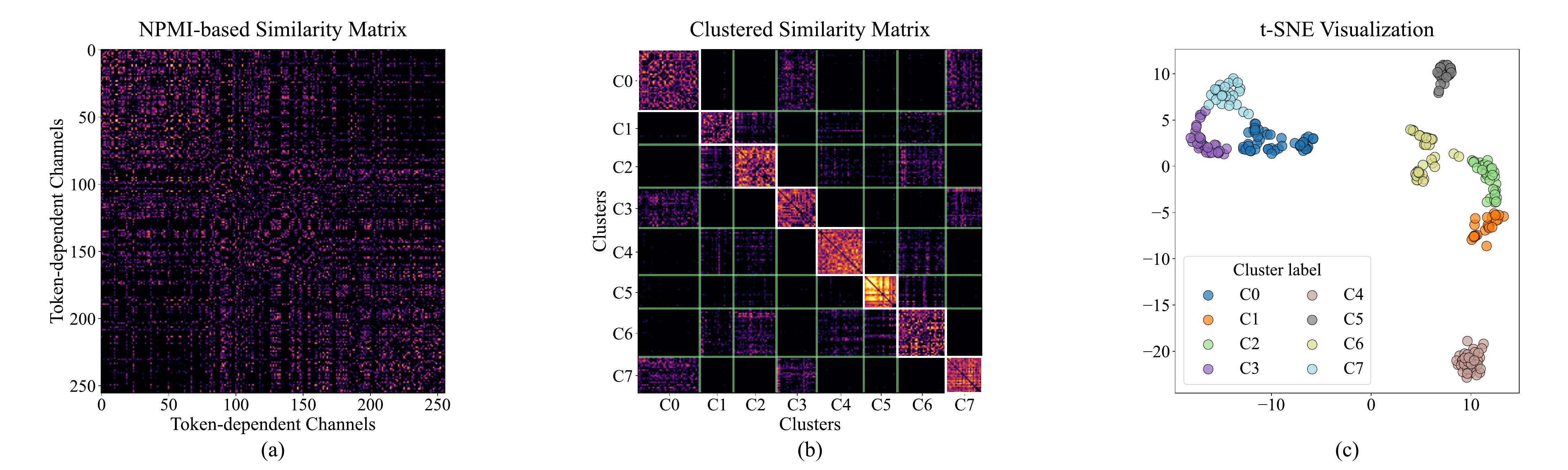}
\vspace{-20pt}
\caption{Illustration of the Co-Occurrence-Based Clustering in a Transformer Block of Qwen2VL-2B.
(a) Similarity matrix $\mathbf{S}^l$ showing mutual co-occurrence among token-dependent channels, with brightness indicating similarity.
(b) Channels with strong co-occurrence are grouped into the same cluster.
(c) t-SNE~\cite{tsne} projection demonstrates that the clustering effectively captures their co-occurrence relations.}\label{fig:cls}
\end{figure*}

\noindent\textbf{Effect of Each Component.} 
Our method integrates a shared expert (SE) and routed experts (REs).
Main results show that their combination yields notable gains across quantization settings.
To further verify each expert type, we conduct ablation studies in~\cref{tab:ablation}. 
Results show that removing either expert consistently degrades performance.
The \emph{random routing} experiment demonstrates that the proposed routing method can adaptively select the suitable routed expert to recover model accuracy.
Similarly, the \emph{random clustering} experiment confirms that the proposed co-occurrence-based clustering substantially enhances quantization performance.
Furthermore, we visualize the clustering results in~\cref{fig:cls} according to \cref{sec:method:res}.
The clustered similarity matrix and t-SNE~\cite{tsne} projection indicate that our method effectively identifies and partitions these co-occurring clusters.

\begin{table}[h]
\centering
\resizebox{0.9\linewidth}{!}{
\begin{tabular}{c|l|cc}
\toprule
Setting & Component & MMMU ($\uparrow$) & ScienceQA ($\uparrow$) \\
\midrule
FP16 & - & 39.89 & 76.95 \\
\midrule
\multirow{5}{*}{W4A6} 
& routed experts (REs)     & 34.56 & 68.72 \\
& shared expert (SE)       & 35.22 & 69.61 \\
& SE + random routing      & 35.89 & 70.00 \\
& SE + random clustering   & 35.33 & 69.71 \\
& \methodabbr{} (SE+REs) & \textbf{36.89} & \textbf{70.85} \\
\midrule
\multirow{5}{*}{W4A8} 
& routed experts (REs)     & 36.00 & 71.94 \\
& shared expert (SE)       & 36.78 & 73.13 \\
& SE + random routing      & 37.89 & 73.67 \\
& SE + random clustering   & 37.22 & 73.82 \\
& \methodabbr{} (SE+REs) & \textbf{38.00} & \textbf{74.37} \\
\bottomrule
\end{tabular}
}
\vspace{-5pt}
\caption{Ablation study results on Qwen2VL-2B model.}\label{tab:ablation}
\vspace{-5pt}
\end{table}

\noindent\textbf{Effect of Refinement.}
\cref{tab:ft}~presents the ablation results of the proposed refinement for routed experts under the Qwen2VL W4A6 quantization.
As observed on both 2B and 7B models, applying refinement consistently leads to notable accuracy improvements across multiple tasks compared to the non-refinement counterparts.

\begin{table}[h]
\centering
\resizebox{0.9\linewidth}{!}{
\begin{tabular}{c|c|ccccc}
\toprule
Model & Ref. & MMMU & OCRBench & ScienceQA & TextVQA & VizWiz \\
\midrule
\multirow{2}{*}{2B} & \ding{55} & 33.78 & 68.20 & \textbf{71.84} & 73.18 & 59.62\\
~& \ding{51} & \textbf{36.89} & \textbf{69.60} & 70.85 & \textbf{73.30} & \textbf{60.58}\\
\midrule
\multirow{2}{*}{7B} & \ding{55} & \textbf{45.44} & 73.00 & 79.87 & 71.63 & 64.18\\
~& \ding{51} & 44.00 & \textbf{74.60} & \textbf{80.61} & \textbf{77.58} & \textbf{65.11}\\
\bottomrule
\end{tabular}
}
\vspace{-5pt}
\caption{Ablation study of Refinement on Routed Experts. ``\ding{51}'' indicates applied refinement; ``\ding{55}'' indicates none.}\label{tab:ft}
\vspace{-5pt}
\end{table}

\noindent\textbf{Effect of the Number of Routed Experts.}
We analyze in \cref{tab:ne} the impact of the number of routed experts on model accuracy.
Here, $N_r$ denotes the number of routed experts. 
It can be observed that as $N_r$ increases, the model performance gradually improves; however, a larger $N_r$ also implies higher memory overhead for routed experts.

\begin{table}[h]
\centering
\resizebox{0.75\linewidth}{!}{
\begin{tabular}{c|ccc|c}
\toprule
$N_r$ & OCRBench & TextVQA & VizWiz & Avg. ($\uparrow$) \\

\midrule
2 & 68.40 & 73.14 & 59.70 & 67.08\\
4 & 68.50 & 73.13 & 60.41 & 67.35\\
8 & 69.60 & 73.30 & 60.58 & 67.83\\
16 & 69.90 & 73.52 & 60.75 & 68.06\\
\bottomrule
\end{tabular}
}
\vspace{-5pt}
\caption{Impact of the number of routed experts on the performance of Qwen2VL-2B under the W4A6 quantization setting.}\label{tab:ne}
\vspace{-5pt}
\end{table}

\subsection{Overheads Analysis and Kernel Performance}

The low-rank adapter introduces additional computation and memory overhead from the lightweight auxiliary matrices $\mathbf{L}_{\cdot A}$ and $\mathbf{L}_{\cdot B}$. 
Let $s$ denote the sequence length, $d$ hidden size, $r(\ll d)$ the rank, and $N_r(\ll d)$ the routed experts count. 
The complexity analysis of a single layer is provided in \cref{tab:overhead}, with computation quantified in terms of floating-point operations (FLOPs) and memory overhead evaluated by the total number of parameters.
Notably, \methodabbr{} incurs only minimal additional computational and memory costs compared to the original linear, yet enables the quantized model to recover accuracy comparable to full precision.

\begin{table}[th]
\centering
\resizebox{0.75\linewidth}{!}{
\begin{tabular}{lcc}
\toprule
Complexity & Origin & \methodabbr\ \\
\midrule
Computation & $sd^2$ & $sd^2+sd(2r+N_r)$ \\
Memory & $d^2$ & $d^2+rd(1+N_r)$ \\
\bottomrule
\end{tabular}
}
\vspace{-5pt}
\caption{Complexity analysis of the linear layer in \methodabbr{} method.}\label{tab:overhead}
\vspace{-5pt}
\end{table}

To assess hardware efficiency, we develop an analytical performance model following the FlightLLM~\cite{flightllm} accelerator architecture. 
Prefill-stage kernel speedups of linear layers are measured with sequence length 128 using Qwen2VL-7B weight shapes under various quantization settings. 
As shown in \cref{tab:npu}, \methodabbr{} achieves $3.5\times$-$4.5\times$ acceleration, highlighting its strong potential for hardware-level efficiency gains.

\begin{table}[th]
\centering
\resizebox{0.75\linewidth}{!}{
\begin{tabular}{r@{\,$\times$\,}lccc}
\toprule
\multicolumn{2}{c}{Shape (IC $\times$ OC)} & W4A6 & W4A8 & W3A16 \\
\midrule
3584 & 3584  & $3.56\times$ & $3.50\times$ & $4.10\times$ \\
3584 & 18944 & $3.60\times$ & $3.59\times$ & $4.50\times$ \\
18944 & 3584 & $3.84\times$ & $3.77\times$ & $4.50\times$ \\
\bottomrule
\end{tabular}
}
\vspace{-5pt}
\caption{NPU speedup ratios of \methodabbr{} for Qwen2VL-7B linear layers compared with the fp16 model, measured during the prefill stage with a sequence length of $s{=}128$. ``IC'' and ``OC'' denote the input and output channel dimensions, respectively.}\label{tab:npu}
\vspace{-5pt}

\end{table}
\section{Conclusion}
\label{sec:con}

In this work, we reveal a key observation that the distributions and occurrence frequencies of important channels vary significantly both across modalities and among tokens, even within the same modality.
Building on this insight, we propose~\methodfull~(\methodabbr{}), a token-aware adaptive error compensation for VLMs quantization, that dynamically adapts to such variations. 
Specifically, \methodabbr{} employs a shared expert to robustly reconstruct token-independent channels and routed experts to adaptively compensate token-dependent ones, with each expert implemented as a low-rank adapter.
Comprehensive evaluations on diverse VLMs show that~\methodabbr{} consistently outperforms globally static PTQ baselines across different quantization configurations.
{
\small
\bibliographystyle{ieeenat_fullname}
\bibliography{main}
}

\clearpage
\setcounter{page}{1}
\maketitlesupplementary

In the supplementary material, we provide additional Related Work, Method Details, and Experimental Results.
In \cref{sec:adm}, we present more complete implementation details for the Shared Expert and the Refinement of Routed Experts.
In \cref{sec:ae}, we report the main results of \methodabbr{} on Qwen2VL-7B and InternVL2-2B, along with evaluations on language tasks.
We further present results for joint quantization of the Visual Encoder and the VLM, along with an extended ablation study on the Number of Important Channels.

\section{Additional Details for Related Work}
\label{sec:rw}

In large language model (LLM) compression, two mainstream approaches are commonly used: quantization-aware training (QAT) and post-training quantization (PTQ).
QAT explicitly models quantization errors during training and can achieve higher accuracy for low-bit models, but it incurs substantial computational and data overhead (\eg, LSQ~\cite{lsq}, LLM-QAT~\cite{llmqat}, DL-QAT~\cite{dlqat}).
In contrast, PTQ directly maps pretrained weights and activations into low-bit representations after training, requiring only a small amount of calibration data.
Due to its efficiency and practicality, PTQ has become the dominant solution for resource-constrained scenarios (\eg, QBB~\cite{qbb}, aespa~\cite{aespa}).
However, PTQ inevitably introduces quantization errors, and existing methods remain constrained by limited outlier identification and error compensation, posing a key challenge for advancing low-bit LLM deployment~\cite{2403.06408}.

To address this core challenge, various solutions have been proposed from different perspectives.
OBQ~\cite{obq} and GPTQ~\cite{gptq} perform progressive quantization with Hessian-guided iterative compensation, allowing unquantized parameters to absorb the quantization errors yielded in previous channels or blocks, thereby alleviating reconstruction error within Transformer blocks.
\cite{achieving} further combines Hessian-based optimization with the Expectation-Maximization (EM) algorithm to enable joint weight-activation quantization at extremely low bitwidths.
Distribution reshaping approaches mitigate the effects of outliers by applying channel-wise scaling and equalization to balance the dynamic ranges of activations and weights.
Among them, SmoothQuant~\cite{smoothquant} transfers part of the quantization difficulty from activations to weights through channel-wise scaling, effectively balancing their dynamic ranges.
Furthermore, AWQ~\cite{Awq} employs a search-based channel scaling strategy, while selectively retaining the most sensitive parameters in full precision to preserve model accuracy.
OmniQuant~\cite{omniquant} incorporates learnable clipping and equivalent scaling transformations, jointly optimized under a block-level error minimization framework to achieve stronger error suppression.
From another perspective, some approaches utilize rotation-based transformations to mitigate outliers in weight and activation quantization, effectively reducing quantization errors.
QuIP \cite{quip} employs an uncorrelated transformation combined with adaptive rounding to minimize proxy errors, while QuIP\#{} \cite{nquip} integrates random Hadamard transforms and block-wise vector quantization to improve reconstruction accuracy.
QuaRot~\cite{quarot} further proposes an end-to-end 4-bit quantization scheme based on Hadamard rotation, which enables simultaneous quantization of weights, activations, and KV cache.
In model quantization, performance degradation and quantization errors primarily arise from outlier and sensitivity-prone important channels. Precisely identifying and preserving these channels at higher precision is essential for mitigating quantization errors.
For instance, Atom~\cite{atom} enhances robustness under low-bit settings through hybrid precision and dynamic activation quantization, whereas SpQR~\cite{spqr} leverages Hessian-based sensitivity analysis to identify important parameters, retaining high precision for outlier weights while quantizing the remaining ones into low-bit representations, thereby effectively mitigating outlier-induced errors.
Another research direction introduces low-rank structures into quantization error compensation by attaching lightweight high-precision low-rank modules to recover accuracy with minimal computational and memory overhead. 
Representative approaches include LoRC~\cite{lorc}, which models quantization residuals using low-rank matrices to restore performance at low cost; LQER~\cite{lqer}, which leverages activation statistics and diagonal rescaling for weighted low-rank reconstruction; and ASER~\cite{ASER}, which adopts whitened SVD for more stable error modeling and integrates outlier-channel analysis to smooth activation distributions.

\section{Additional Details for Method}\label{sec:adm}

\subsection{Shared Expert}\label{subsec:se}

The construction process of the shared expert in \methodabbr\ is illustrated in \cref{alg:se}, which follows the general procedure described in~\cite{ASER}.
This method employs a low-rank structure to approximate the quantization error introduced by weight quantization, with a particular focus on frequently activated, token-independent important channels, thereby effectively capturing globally stable quantization error patterns.

\begin{algorithm}[htbp]
\caption{Building the Shared Expert}
\label{alg:se}
\DontPrintSemicolon
\SetKwInOut{Input}{Input}
\SetKwInOut{Output}{Output}
\Input{Per-layer data $\{\mathbf{X}^l,\,\mathbf{W}^l_f,\,\mathcal{C}^l_s\}_{l=1}^{L}$, quantizer $Q(\cdot)$; rank $r$.}

\Output{Quantized layer weight $\{\mathbf{W}_q^l\}_{l=1}^{L}$, SE $\{(\mathbf{L}_{SA}^l, \mathbf{L}_{SB}^l)\}_{l=1}^{L}$, and residual errors $\{\mathbf{E}_S^l\}_{l=1}^{L}$.}
\BlankLine
Compute $\mathbf{x}^l \leftarrow \operatorname{Mean}_{\text{row}}(|\mathbf{X}^l|)$\;
\For{$l\gets 1$ \KwTo $L$}{
    Initialize $\omega=[1,1,\dots,1]^n$, $\Omega=\mathrm{diag}(\omega)$\;
    Compute $\omega_{\mathcal{C}^l_s} = \mathbf{x}^l_{\mathcal{C}^l_s} / \min(\mathbf{x}^l_{\mathcal{C}^l_s})$\;
    $\mathbf{E}_q^l = \mathbf{W}^l_f - Q(\mathbf{W}^l\mathrm{diag}(\mathbf{1} - \mathbf{1}_{\mathcal{C}^l_s}))$\;
    Compute whitening matrix $S$ by Cholesky decomposition of 
    $(\Omega^{-1}X)(\Omega^{-1}X)^\top$ such that 
    $(S^{-1}\Omega^{-1}X)(S^{-1}\Omega^{-1}X)^\top = I$\;
    Perform SVD: $U\Sigma V^\top = E_q^l S$\;
    Compute: $\mathbf{L}_{SA}^l = U_r \Sigma_r$, $\mathbf{L}_{SB}^l = V_r^\top S^{-1}$, $\mathbf{E}_S^l = \mathbf{E}_q^l - \mathbf{L}_{SA}^l \mathbf{L}_{SB}^l$\;
}
\Return{$\{\mathbf{W}_q^l\}_{l=1}^{L}$, $\{(\mathbf{L}_{SA}^l, \mathbf{L}_{SB}^l)\}_{l=1}^{L}$, $\{\mathbf{E}_S^l\}_{l=1}^L$}\;
\end{algorithm}

\subsection{Refinement of Routed Experts}
\label{suppl:ft}

In this subsection, we describe the loss functions used in the Refinement stage.
These losses follow standard formulations commonly adopted in prior research.
We provide detailed explanations here due to space limitations in the main paper.
The refinement objective consists of two complementary losses: a regression loss $\mathcal{L}_{\mathrm{reg}}$ and a classification loss $\mathcal{L}_{\mathrm{cls}}$.
$\mathcal{L}_{\mathrm{reg}}$ aims to minimize the reconstruction error between the quantized output $\hat{y}$ and full-precision output $y$, encouraging each expert to specialize in its own direction of compensation.
$\mathcal{L}_{\mathrm{cls}}$ improves the router’s ability to predict the optimal expert for a given input.

Specifically, let $y_i$ denote the output reconstructed by the $i$-th routed expert and $y$ the full-precision output.
We define the reconstruction distance as
$d_i = \| \hat{y}_i - y\|_1$.
During refinement, only the routed expert achieving the smallest reconstruction error is optimized, formulated as:
\begin{align}
\mathcal{L}_{\mathrm{reg}} &= \min_{i \in [1, N_r]} {d_i}.
\end{align}

To enable the router to predict the relative performance of different routed experts, we denote its output as $l = R|x|$ and construct a classification objective based on the inter-expert discrepancy.
We adopt the Kullback-Leibler divergence to align the predicted distribution with the normalized reconstruction loss distribution:
\begin{align}
\mathcal{L}_{\mathrm{cls}} &= \tau^2 D_{\mathrm{KL}}\left(\mathbf{P} \parallel \mathbf{Q} \right), \\
\mathbf{P} &= \mathrm{softmax}\left(\frac{-(d - \mu(d))/{\sigma(d)}}{\tau}\right), \\
\mathbf{Q} &= \mathrm{softmax}\left(\frac{-(l - \mu(l))}{\tau}\right),
\end{align}
where $\tau$ is a temperature coefficient, and $\mu(\cdot)$ and $\sigma(\cdot)$ denote the mean and standard deviation.
Finally, we use two coefficients, $\alpha$ and $\beta$, to balance the two losses:
\begin{align}
\mathcal{L}=\alpha\mathcal{L}_{\mathrm{reg}}+\beta\mathcal{L}_{\mathrm{cls}}.
\end{align}

\section{Additional Experiments}\label{sec:ae}

\begin{table*}[htbp]
  \centering
  \resizebox{1.0\textwidth}{!}{
      \begin{tabular}{l|cc|ccccccccccc|c}
        \toprule
        Method & \#W & \#A & MMMU & OCRBench & ScienceQA & TextVQA & VizWiz & AI2D & ChartQA & DocVQA & InfoVQA & MMStar & MuriBench & Avg. ($\uparrow$) \\
        \midrule
    
        Qwen2VL-7B & 16 & 16 & 50.78 & 79.50 & 84.83 & 81.48 & 68.56 & 80.60 & 81.68 & 91.68 & 69.77 & 57.74 & 42.92 & 71.78 \\
        \midrule
        \midrule
        
        RTN & 4 & 6 & 39.00 & 59.50 & 75.41 & 66.93 & 56.64 & 69.62 & 72.00 & 78.73 & 52.91 & 49.97 & 38.27 & 59.91\\
        SQ (ICML'23) & 4 & 6 & 41.00 & 63.90 & 77.09 & 67.94 & 57.48 & 70.92 & 68.84 & 78.78 & 52.90 & 50.48 & 35.65 & 60.45\\
        LQER (ICML'24) & 4 & 6 & 42.56 & 65.60 & 77.24 & \textbf{71.87} & \textbf{64.44} & 71.44 & 74.04 & 81.57 & 56.86 & 49.75 & 41.85 & 63.38\\
        MBQ (CVPR'25) & 4 & 6 & 40.56 & 62.70 & 79.67 & 70.91 & 51.48 & 71.31 & 72.20 & 81.99 & 54.71 & 47.37 & 34.54 & 60.68\\
        \midrule
        \methodabbr{} & 4 & 6 & \textbf{45.44} & \textbf{73.00} & \textbf{79.87} & 71.63 & 64.18 & \textbf{75.58} & \textbf{74.16} & \textbf{83.60} & \textbf{60.45} & \textbf{52.63} & \textbf{42.19} & \textbf{65.70}\\

        \midrule
        \midrule
        
        RTN & 4 & 8 & 45.44 & 60.30 & 79.47 & 71.18 & 59.11 & 76.78 & 74.52 & 77.04 & 56.89 & 53.23 & 40.04 & 63.09\\
        SQ (ICML'23) & 4 & 8 & 43.78 & 58.60 & 79.52 & 69.52 & 53.01 & 76.20 & 72.88 & 74.34 & 53.95 & 52.93 & 36.69 & 61.04\\
        LQER (ICML'24) & 4 & 8 & \textbf{48.00} & 69.50 & \textbf{81.76} & 75.31 & 66.02 & 77.85 & 77.04 & 82.27 & 61.52 & \textbf{55.24} & \textbf{43.85} & 67.12\\
        MBQ (CVPR'25) & 4 & 8 & 46.33 & 72.00 & 81.46 & 75.35 & 60.78 & 76.98 & 76.32 & 85.26 & 61.62 & 53.52 & 37.65 & 66.12\\
        \midrule
        \methodabbr{} & 4 & 8 & 46.33 & \textbf{78.20} & 81.51 & \textbf{78.98} & \textbf{66.59} & \textbf{79.05} & \textbf{78.92} & \textbf{89.21} & \textbf{66.16} & 54.04 & 42.46 & \textbf{69.22}\\
        
        \midrule
        \midrule
        
        RTN & 3 & 16 & 32.44 & 65.80 & 67.63 & 70.43 & 55.81 & 76.75 & 73.72 & 74.46 & 58.14 & 50.32 & 43.69 & 60.84\\
        AWQ (MLSys'24) & 3 & 16 & \textbf{48.00} & 76.30 & 82.15 & 79.00 & 65.69 & 76.49 & 78.92 & 89.10 & 64.98 & \textbf{54.01} & 40.69 & 68.67\\
        LQER (ICML'24) & 3 & 16 & 46.44 & 64.50 & 80.81 & 72.50 & 66.22 & 76.91 & 75.36 & 77.42 & 59.77 & 52.01 & 43.27 & 65.02\\
        MBQ (CVPR'25) & 3 & 16 & 46.22 & 74.40 & \textbf{82.35} & 79.43 & 65.02 & 77.30 & 78.24 & 88.59 & 64.70 & 52.45 & 42.96 & 68.33\\
        \midrule
        \methodabbr{} & 3 & 16 & 46.67 & \textbf{77.20} & 81.56 & \textbf{79.87} & \textbf{67.20} & \textbf{78.01} & \textbf{79.28} & \textbf{89.60} & \textbf{65.26} & 53.35 & \textbf{44.58} & \textbf{69.33}\\

        \bottomrule
      \end{tabular}
    }
  \caption{Main results on the model of Qwen2VL-7B.}\label{tab:qwen7b}
  \vspace{-10pt}
\end{table*}

\begin{table*}[htbp]
  \centering
  \resizebox{1.0\textwidth}{!}{
      \begin{tabular}{l|cc|ccccccccccc|c}
        \toprule
        Method & \#W & \#A & MMMU & OCRBench & ScienceQA & TextVQA & VizWiz & AI2D & ChartQA & DocVQA & InfoVQA & MMStar & MuriBench & Avg. ($\uparrow$) \\
        \midrule
    
        InternVL2-2B & 16 & 16 & 34.33 & 75.30 & 94.30 & 72.58 & 45.94 & 72.83 & 74.84 & 84.84 & 53.23 & 48.20 & 28.46 & 62.26 \\
        \midrule
        \midrule
        
        RTN & 4 & 6 & 30.44 & 67.10 & 86.47 & 66.17 & 41.66 & 63.41 & 66.48 & 77.41 & 43.91 & 40.11 & 26.85 & 55.46\\
        SQ (ICML'23) & 4 & 6 & 31.89 & 69.30 & 88.25 & 67.24 & 40.17 & 64.41 & 65.48 & 79.94 & 46.94 & 41.62 & 25.69 & 56.45\\
        LQER (ICML'24) & 4 & 6 & 30.78 & 70.90 & 88.35 & 67.81 & 39.33 & 65.64 & 68.92 & 79.78 & 45.73 & 41.75 & 28.27 & 57.02\\
        MBQ (CVPR'25) & 4 & 6 & 31.33 & 70.90 & 90.53 & 68.54 & 41.39 & 67.52 & 70.20 & 80.99 & 47.95 & \textbf{45.26} & 25.46 & 58.19\\
        \midrule
        \methodabbr{} & 4 & 6 & \textbf{32.11} & \textbf{72.80} & \textbf{92.12} & \textbf{70.41} & \textbf{43.81} & \textbf{68.69} & \textbf{70.52} & \textbf{82.00} & \textbf{48.69} & 45.18 & \textbf{28.69} & \textbf{59.55}\\

        \midrule
        \midrule
        
        RTN & 4 & 8 & 32.00 & 72.10 & 91.08 & 69.19 & 42.72 & 68.07 & 69.04 & 81.06 & 48.97 & 45.12 & 28.27 & 58.87\\
        SQ (ICML'23) & 4 & 8 & 33.78 & 71.20 & 91.27 & 69.20 & 40.19 & 68.13 & 68.04 & 81.31 & 48.75 & 44.23 & 26.69 & 58.44\\
        LQER (ICML'24) & 4 & 8 & \textbf{34.56} & 72.50 & 92.07 & 70.53 & 39.16 & 69.33 & 70.96 & 82.03 & 49.89 & 44.84 & 28.65 & 59.50\\
        MBQ (CVPR'25) & 4 & 8 & 32.78 & 72.50 & 92.22 & 70.19 & \textbf{44.31} & 70.53 & 71.44 & 82.24 & 49.89 & \textbf{47.73} & 27.31 & 60.10\\
        \midrule
        \methodabbr{} & 4 & 8 & 32.33 & \textbf{74.00} & \textbf{92.86} & \textbf{71.44} & 43.29 & \textbf{71.47} & \textbf{72.88} & \textbf{83.30} & \textbf{51.11} & 45.77 & \textbf{29.08} & \textbf{60.68}\\

        \midrule
        \midrule
        
        RTN & 3 & 16 & 29.78 & 69.70 & 88.65 & 67.51 & 38.21 & 66.09 & 68.88 & 80.56 & 46.29 & 41.45 & 28.38 & 56.86\\
        AWQ (MLSys'24) & 3 & 16 & 29.78 & 69.50 & 89.89 & 68.12 & 45.30 & 67.78 & 68.44 & 80.84 & 46.44 & 44.68 & 25.50 & 57.84\\
        LQER (ICML'24) & 3 & 16 & \textbf{31.00} & 70.30 & 89.34 & 67.80 & 35.93 & 67.03 & 69.52 & 80.41 & 46.57 & 41.74 & 28.04 & 57.06\\
        MBQ (CVPR'25) & 3 & 16 & 30.33 & 69.20 & 89.39 & 67.83 & 45.74 & 67.68 & 68.40 & 80.57 & 46.21 & 44.20 & 26.27 & 57.80\\
        \midrule
        \methodabbr{} & 3 & 16 & 30.78 & \textbf{72.10} & \textbf{92.76} & \textbf{69.60} & \textbf{47.68} & \textbf{70.05} & \textbf{71.44} & \textbf{82.16} & \textbf{47.90} & \textbf{45.08} & \textbf{29.77} & \textbf{59.94}\\

        \bottomrule
      \end{tabular}
    }
  \caption{Main results on the model of InternVL2-2B.}\label{tab:internvl2b}
  \vspace{-10pt}
\end{table*}

\subsection{Additional Model}

The experimental results of \methodabbr\ on additional models are presented in \cref{tab:qwen7b} and \cref{tab:internvl2b}. 
Consistent with previous findings, our method significantly outperforms the baselines under W4A6, W4A8, and W3A16 configurations.

\subsection{Performance on Language Tasks}

The core idea of \methodabbr\ is to employ multi-expert low-rank adapters that dynamically adapt to compensation differences across modalities and even among individual tokens, thereby improving model performance on both vision-language and language-only tasks.
To validate this, we evaluate the quantized Qwen2VL-2B and Qwen2VL-7B models on the MMLU benchmark under different quantization methods.
As shown in \cref{tab:llm}, \methodabbr\ consistently achieves significant performance gains over LQER across various quantization configurations and model scales.
These results demonstrate that explicitly modeling sensitivity differences across modalities and tokens not only effectively mitigates performance degradation in vision-language tasks but also helps maintain stable performance on language-only tasks.

\begin{table}[h]
\centering
\resizebox{0.75\linewidth}{!}{
\begin{tabular}{cccc}
\toprule
Model & Setting & Method & MMLU ($\uparrow$) \\
\midrule
\multirow{5}{*}{Qwen2VL-2B} & FP16 & - & 52.79 \\
\cmidrule(){2-4}
~ & \multirow{2}{*}{W4A6}  & LQER & 44.37 \\
~ & ~ & QE   & \textbf{47.21} \\
\cmidrule(){2-4}
~ & \multirow{2}{*}{W4A8}  & LQER & 46.60 \\
~ & ~ & QE   & \textbf{50.35} \\
\midrule
\multirow{5}{*}{Qwen2VL-7B} & FP16 & - & 67.88 \\
\cmidrule(){2-4}
~ & \multirow{2}{*}{W4A6}  & LQER & 55.59 \\
~ & ~ & QE   & \textbf{61.87} \\
\cmidrule(){2-4}
~ & \multirow{2}{*}{W4A8}  & LQER & 64.21 \\
~ & ~ & QE   & \textbf{64.83} \\
\bottomrule
\end{tabular}
}
\caption{The results of quantized Qwen2VL on the MMLU benchmark.}\label{tab:llm}\vspace{-10pt}
\end{table}

\subsection{Quantize Both Visual Encoder and VLM}

To achieve higher acceleration ratios, we further quantize both the visual encoder and the merger module that connects the encoder to the VLM.
As shown in \cref{tab:vlm_quantization}, L denotes the VLM, V the visual encoder, and M the merger module, where \ding{51} indicates that the corresponding module is quantized.
The results show that as more modules are quantized, the model exhibits negligible performance degradation, demonstrating that the proposed joint quantization strategy effectively improves overall efficiency while maintaining accuracy.

\begin{table}[th]
\centering
\resizebox{1.0\linewidth}{!}{
\begin{tabular}{ccc|c c c c|c}
\toprule
L & V & M & OCRBench & ScienceQA & TextVQA & VizWiz & Avg. ($\uparrow$) \\
\midrule
-  & -  & - & 74.90 & 76.95 & 77.72 & 65.73 & 73.83 \\
\midrule
\ding{51}  & -  & - & 68.20 & 71.84 & 73.18 & 59.62 & 68.21 \\
\ding{51}  & \ding{51}  & - & 65.90 & 70.75 & 72.09 & 59.83 & 67.14 \\
\ding{51}  & \ding{51} & \ding{51} & 66.40 & 70.10 & 71.77 & 59.28 & 66.89 \\
\bottomrule
\end{tabular}
}
\caption{Quantization results of different modules in Qwen2VL-2B. 
The symbol ``-‘’ indicates full precision (FP16), while \ding{51} denotes W4A6 quantization. 
L, V, and M correspond to the VLM, visual encoder, and merger module, respectively.}\label{tab:vlm_quantization}
\vspace{-10pt}
\end{table}

\subsection{Effect of the Number of Important Channels}

We further investigate the effect of the number of important channels $k$ on model accuracy in \cref{tab:k}. The results indicate a steady improvement as $k$ increases. However, at $k=64$, the performance saturates and slightly declines, as selecting an excessively large set of channels dilutes the focus on truly critical ones.

\begin{table}[h]
\centering
\resizebox{0.75\linewidth}{!}{
\begin{tabular}{c|ccc|c}
\toprule
$k$ & MMMU & OCRBench & VizWiz & Avg. ($\uparrow$) \\
\midrule
4  & 34.89 & 67.30 & 58.93 & 53.71 \\
8  & 35.33 & 68.90 & 58.95 & 54.39 \\
16 & 35.11 & 68.50 & 59.92 & 54.51 \\
32 & 36.11 & 69.30 & 60.24 & 55.22 \\
64 & 34.44 & 70.10 & 59.57 & 54.70 \\
\bottomrule
\end{tabular}
}
\caption{Impact of the number of important channels on the performance of Qwen2VL-2B under the W4A6 quantization setting.}\label{tab:k}
\vspace{-10pt}
\end{table}
    
\end{CJK}
\end{document}